\documentclass[conference]{IEEEtran}
\usepackage{times}

\usepackage{multicol}
\usepackage[bookmarks=true,colorlinks=true,citecolor=blue,linkcolor=blue]{hyperref}
\usepackage{microtype}
\usepackage{amsmath}%
\usepackage{amssymb}%
\usepackage{graphics}%
\usepackage[tight]{subfigure}
\usepackage{tabularx}
\usepackage{booktabs}
\usepackage{multirow}
\usepackage{xspace} %
\usepackage{flushend}
\usepackage[ruled]{algorithm2e}
\usepackage{bm}
\usepackage{doi}
\usepackage{algorithmic}

\usepackage{acronym}
\acrodef{MDP}{Markov Decision Process}
\acrodef{POMDP}{Partially Observable Markov Decision Process}
\acrodef{DNN}{deep neural network}
\acrodef{CNN}{convolutional neural network}
\acrodef{RNN}{recursive neural network}
\acrodef{LSTM}{long short-term memory}
\acrodef{IRL}{inverse reinforcement learning}
\acrodef{IOC}{inverse optimal control}

\usepackage[backend=biber,natbib=true,style=alphabetic,citestyle=numeric,sorting=nty,maxbibnames=99,maxcitenames=1,mincitenames=1,bibstyle=ieee,dateabbrev=false]{biblatex}
\DefineBibliographyStrings{english}{
  andothers   = {et al\adddot}\addspace
}

\DeclareSortingTemplate{ymdt}{
  \sort{
    \field{presort}
  }
  \sort[final]{
    \field{sortkey}
  }
  \sort[direction=descending]{
    \field{sortyear}
    \field{year}
    \literal{9999}
  }
  \sort[direction=descending]{
    \field[padside=left,padwidth=2,padchar=0]{month}
    \literal{99}
  }
  \sort{
    \field{sorttitle}
  }
  \sort[direction=descending]{
    \field[padside=left,padwidth=4,padchar=0]{volume}
    \literal{9999}
  }
}

\usepackage{color}
\usepackage[textsize=scriptsize]{todonotes} %
\usepackage{url}
\usepackage{subfiles}

\begin{document}

\title{Soft Robots Learn to Crawl: Jointly Optimizing Design and Control with Sim-to-Real Transfer}

\author{\IEEEauthorblockN{Charles Schaff\IEEEauthorrefmark{1}, Audrey Sedal\IEEEauthorrefmark{2}, Matthew R.~Walter\IEEEauthorrefmark{1}} \IEEEauthorblockA{\IEEEauthorrefmark{1}Toyota Technological Institute at Chicago, Illinois, USA 60637\\
Email: \texttt{\{cbschaff,mwalter\}@ttic.edu}} \IEEEauthorblockA{\IEEEauthorrefmark{2}Department of Mechanical Engineering, McGill University, Montreal, Canada\\
Email: \texttt{audrey.sedal@mcgill.ca}}}

\maketitle

\begin{abstract}
This work provides a complete framework for the simulation, co-optimization, and sim-to-real transfer of the design and control of soft legged robots. The compliance of soft robots provides a form of ``mechanical intelligence''---the ability to passively exhibit behaviors that would otherwise be difficult to program. Exploiting this capacity requires careful consideration of the  coupling between mechanical design and control.
Co-optimization provides a promising means to generate sophisticated soft robots by reasoning over this coupling.
However, the complex nature of soft robot dynamics makes it difficult to provide a simulation environment that is both sufficiently accurate to allow for sim-to-real transfer, while also being fast enough for contemporary co-optimization algorithms.
In this work, we show that finite element simulation combined with recent model order reduction techniques provide both the efficiency and the accuracy required to successfully learn effective soft robot design-control pairs that transfer to reality. We propose a reinforcement learning-based framework for co-optimization and demonstrate successful optimization, construction, and zero-shot sim-to-real transfer of several soft crawling robots.
Our learned robot outperforms an expert-designed crawling robot, showing that our approach can generate novel, high-performing designs even in well-understood domains.
\end{abstract}

\IEEEpeerreviewmaketitle

\section{Introduction} \label{sec:intro}

The deformable nature soft robots enables designs that respond to contact or control inputs in sophisticated ways, with behaviors that have proven effective across a variety of domains. %
The design of such systems is tightly coupled with the policy that controls their motion, giving rise to a form of ``mechanical intelligence''~\cite{rus2015design} in which materials and mechanisms respond to their environment in useful ways that augment functionality, e.g., conforming to an object to create a better grasp or storing elastic energy to improve the efficiency and power of a walking gait. Therefore, methods that jointly optimize both the robot's physical design and its control policy provide a promising approach to discovering mechanically intelligent soft robots.
However, while researchers have extensively explored the problem of joint design-control optimization in the context of rigid robots~\cite{sims1994, park94, paul01, paul06, spielberg_functional_2017, seo2019modular, digumarti_concurrent_2014, ha17, zhao2020robogrammar, schaff_jointly_2019, ha_reinforcement_2019, chen_hardware_2020, pathak_learning_nodate}, relatively little work exists for soft robotics.

This work provides a complete framework for the simulation, co-optimization, and sim-to-real transfer of the design and control of soft robots. Integral to this framework, we propose a co-optimization algorithm %
that utilizes multi-task deep reinforcement learning to generate a design-aware policy capable of generalizing across the space of designs. The algorithm exploits this policy to quickly focus its search on high-performing designs. 
To encourage ``mechanical intelligence'', we learn an open-loop controller, forcing complex behavior to be expressed through the resulting soft body.

An important prerequisite for co-optimization is a simulator that is both fast enough to explore a large set of designs and control strategies and accurate enough to ensure that the learned robots are physically realizable and capable of sim-to-real transfer.
However, modelling soft bodies is both challenging and computationally intensive.
The best way to simulate soft bodies for robotics is an open question, and the few co-optimization approaches for soft robotics suggest different simulation strategies~\cite{Hu19,spielberg_co-learning_2021,Hiller2014}.
However, these simulators have varying degrees of realism and their ability to produce soft robots that cross the reality gap is unclear.
Therefore, in addition to designing a co-optimization algorithm, new simulation approaches are also required to improve computational efficiency and transferability to the real world.

\begin{figure}[!t]
    \centering
    \includegraphics[width=\linewidth]{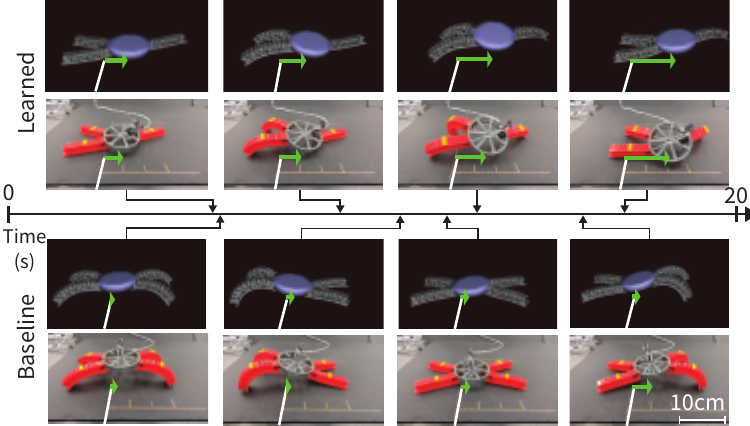}
    \caption{Our framework jointly learns the design and control of crawling soft robots (top) that outperform an expert-designed baseline (bottom). While trained exclusively in simulation, our learned robots are capable of zero-shot sim-to-real transfer, with the optimal design moving more than $2 \times$ faster than the baseline in the real world.}\label{fig:frames}%
\end{figure}
We employ finite element analysis (FEA), which is the defacto standard for simulating deformable materials with a high degree of accuracy. In order to improve the computational complexity of FEA-based simualation while preserving its accuracy, we extend the recent work of \citet{Goury18} that proposes a model order reduction technique for soft robotics in the open-source FEA simulation framework SOFA~\cite{Faure2012, Coevoet17}. Their reduction technique has a large initial computational cost, but then allows for simulating a fixed robot with a computational efficiently sufficient for learning-based methods, while maintaining physical realism.
However, co-optimization requires a search over many unique designs and reducing each one is computationally infeasible.
To overcome this, we propose a reconfigurable reduction framework that reduces a set of composable parts that can then be combined to create reduced order models of soft robots with varying morphologies.

While our approach is general, we focus our study around the easily manufacturable PneuNet actuator~\cite{mosadegh2014pneumatic}, which has previously been used to create robots capable of walking and crawling gaits~\cite{gamus2020understanding, shepherd_multigait_2011}.
We experimentally validate our proposed approach by learning combinations of PneuNets and their controllers that together lead to faster gaits, and demonstrate the ability to successfully transfer optimized design-control pairs to reality.

Our work contributes a complete simulation and optimization framework for the joint design and control of soft robots capable of zero-shot sim-to-real transfer. Specifically, this includes:
\begin{enumerate}
    \item a model-free algorithm for optimizing the blended design and control spaces of soft robots;
    \item a framework for creating reconfigurable reduced-order soft robot models that improve computational efficiency and enable the use of learning techniques;%
    \item the discovery of pneumatically actuated soft robots that outperform a standard expert-designed crawling robot in simulation and reality.
\end{enumerate}
See our webpage\footnote{\href{https://sites.google.com/ttic.edu/evolving-soft-robots}{https://sites.google.com/ttic.edu/evolving-soft-robots}} for code and videos of our results. %

\section{Related Work} \label{sec:related-work}

The problem of jointly optimizing a rigid robot's physical structure along with its control has a long history in robotics research. Early work employs evolutionary methods to optimize the robot's design along with its (often neural) controller~\cite{lipson00, paul01, murata2007self, bongard11}. Another common approach is to assume access to a parameterized model of the robot's dynamics and to then optimize these parameters together with those of control (or motion)~\cite{paul06, villarreal13, ha17, spielberg_functional_2017, geilinger18, taylor2019optimal, bravo2020one}. Bolstered by the availability of efficient high-fidelity physical simulators, joint optimization methods based on reinforcement
learning are able to learn capable rigid-body design-controller pairs without prior knowledge of the dynamics~\cite{schaff_jointly_2019, pathak_learning_nodate, ha_reinforcement_2019, whitman2021learning}.

Compared to rigid robotics, jointly optimizing the design and control of soft robots is less explored. Of the work that exists, the large majority focus exclusively on simulation. Many approaches reason over design and control spaces that include a mix of discrete and continuous parameters (e.g.,
voxel-based soft robots (VSRs)~\cite{talamini2019evolutionary} are composed of discrete voxels, but the input frequency to each voxel is considered to be continuous). %
\citet{spielberg_co-learning_2021} propose an autoencoder-based method that is able to optimize the placement of a large number of such voxels for simulated locomotion tasks, with fewer iterations than other approaches.
\citet{cheney_unshackling_2013} use an evolutionary neural strategy to develop designs for VSRs that locomote in simulation.
\citet{kriegman2019automated} describe an approach to deforming the structure of VSRs subject to damage such that the original control policy remains valid. \citet{ma2021diffaqua} use a material point method-based simulation and gradient-based optimization methods to co-optimize the shape and control of simulated swimming robots. \citet{deimel2017automated} use particle filter-based optimization to co-optimize finger angles and the grasp strategy of a soft gripper.
The success of these methods in simulation is encouraging for soft roboticists, and recent simulation-based benchmarks allow for a rigorous comparison of co-optimization methods~\cite{collins2021review,bhatia2021evolution}. However, existing work provides a limited evaluation of the physical design-control pairs, and so little is known about their ability to transfer to the real world. Indeed, experiments on voxel-based soft robots reveal that their behavior in simulation can differ significantly from reality~\cite{kriegman2020scalable}. 

One notable exception, \citet{morzadec2019toward} experimentally verify an optimized soft robotic joint, showing how shape optimized using a finite element analysis-based simulator~\cite{Coevoet17} translates to improvements in a real-world soft robotic leg, however they do not consider optimizing the controller. Another exception is recent work that integrates a pneumatic-based passive controller into the robot's design to achieve a forward walking gait~\cite{drotman_electronics-free_2021}, providing an example of how soft robots can have unclear boundaries between design and control.

Meanwhile, individual design and control methods continue to be key areas of research in soft robotics~\cite{rus2015design}. 
There exist a wide variety of design concepts for soft robots~\cite{chen2020design}
such as fluidically pressurizeable soft devices~\cite{mosadegh2014pneumatic,shepherd_multigait_2011},
metamaterial-based designs~\cite{rafsanjani2018kirigami, lipton2018handedness}, and cable-driven
devices~\cite{bern2020soft}.
Soft roboticists note that existing design optimization methods for compliant, nonlinear mechanisms, such as topology optimization, are challenging to use in soft robotics due to complicated soft material behavior~\cite{chen2020design}. %
The diversity of the design space for soft robots further exacerbates the challenge of automating the search for optimal designs~\cite{pinskier2021bioinspiration}. 
Model-\cite{bern2019trajectory, bruder2019modeling} and learning-based~\cite{lee2020generalized, Culha-RSS-20, kim2021review} controllers have also proven successful, as well as hybrid policy designs~\cite{vitanov2020shape, bern2020soft, howison_reality-assisted_2020}.
\citet{zhu2019learning} consider an origami-like robot with various design configurations that all inform policy optimization, and \citet{morimoto2021model} employ the soft actor-critic algorithm~\cite{haarnoja2018soft} for reaching tasks. Related, \citet{vikas2016design} present a modular approach to designing 3D-printed motor-tendon soft robots that can be readily fabricated, and a model-free algorithm for learning the corresponding control policy. Unlike our framework, however, they do not jointly reason over design and control.

\section{Co-Optimization of Design and Control} \label{sec:coopt}

\begin{figure}[!t]
    \centering
    \includegraphics[width=\columnwidth]{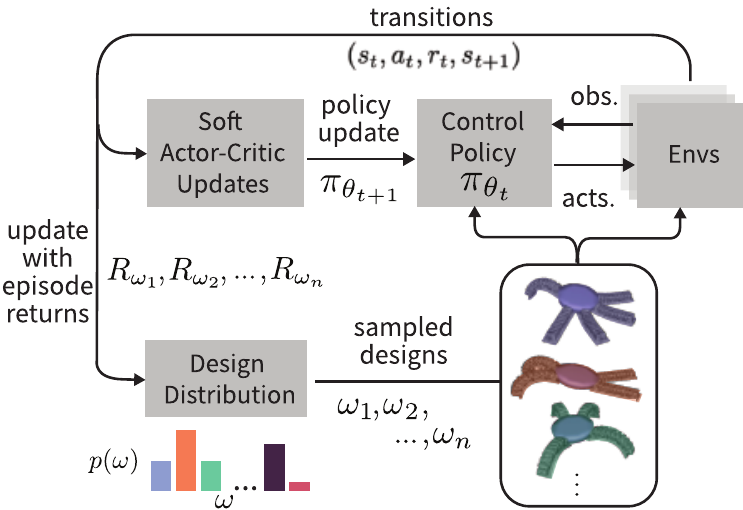}
    \caption{Our approach maintains a distribution over designs $p(\omega)$. At each iteration, the method samples a set of designs $\omega_1, \hdots, \omega_n$ and controls each using a shared, design-conditioned, policy $\pi_\theta$. We train the policy using soft actor-critic on a mixture of data from different designs, and update the design distribution based on the episode returns of the sampled designs.}\label{fig:algorithm}
\end{figure}

We first describe the general approach to jointly optimizing robot design and control, and then discuss a specific application to crawling soft robots.
Algorithm \ref{alg:main} and Figure~\ref{fig:algorithm} give an overview of this approach.

\subsection{General Approach via Multi-task Reinforcement Learning}

\newcommand{\mS}{\mathcal{S}}
\newcommand{\mA}{\mathcal{A}}
\newcommand{\mP}{\mathcal{P}}
\newcommand{\mR}{\mathcal{R}}
\newcommand{\mE}{\mathcal{E}}
\newcommand{\mM}{\mathcal{M}}

The control problem can be modelled as a Markov decision process (MDP) $\mM = \textrm{MDP}(\mS, \mA, \mP, \mR)$, where $\mS$ is the state space, $\mA$ is the action space, $\mP: \mS \times \mA \times \mS \rightarrow [0,1]$ is the transition dynamics, and $\mR: \mS \times \mA \times \mS \rightarrow \mathbb{R}$ is the reward function.
When co-optimizing design and control, we additionally define the design space $\Omega$.
Assuming we are optimizing for a single task specified by its reward $\mR$, we define the design-specific MDP $\mM_\omega = \textrm{MDP}(\mS_\omega, \mA_\omega, \mP_\omega, \mR)$ for each design $\omega \in \Omega$.
In most co-optimization settings with a single task, the state and action spaces will change between designs only when those designs have different morphologies.

Let $\pi^*_\omega: \mS_\omega \times \mA_\omega \rightarrow [0, 1]$ be the optimal policy for MDP $\mM_\omega$. The goal of co-optimization is to find the optimal design and controller pair $(\omega^*$, $\pi^*_{\omega^*})$ such that:
\begin{equation}
    \label{eqn:objective}
    \omega^*, \pi^*_{\omega^*}  = \underset{\omega, \pi_\omega}{\text{arg max}} \; \mathbb{E}_{\pi_\omega}\left[\sum_t \gamma^t \mR_t \right]
\end{equation}

In this setting, we are faced with many MDPs that share common structure. Solving each MDP independently is intractable and ignores these similarities, which are critical to generalizing to new designs. %
We draw on insights from multi-task reinforcement learning~\cite{Varghese2020} to more efficiently solve for the optimal design-control pairs (Eqn.~\ref{eqn:objective}) by exploiting this common structure.
Similar to goal-conditioned policies, our approach learns a single \textit{design-conditioned} policy $\pi_\theta: \mS \times \mA \times \Omega \rightarrow [0, 1]$ to control all the designs in $\Omega$ for the specified task.
This idea was proposed in the context of co-optimization~\cite{schaff_jointly_2019} as well as for the sub-problem of controlling a set of designs with different morphologies~\cite{Huang2020, Kurin2021}.
This policy can be trained using any RL algorithm on a mixture of data collected with designs in $\Omega$.

In order to search over designs, we maintain a distribution $p(\omega)$ over the design space $\Omega$.
This distribution generates designs for training the controller and models the belief about which designs are optimal given the current design-conditioned control policy.
At the start of training, $p(\omega)$ should provide a large diversity of designs and then, once the controller has been sufficiently trained, slowly concentrate probability mass around high-performing designs.
The controller can then specialize to an increasingly promising subset of designs until the algorithm converges on a single design and a controller that is then fine-tuned for that design.

The design distribution can be modeled in a number of different ways depending on the nature of the design space. For example, \citet{schaff_jointly_2019} use a mixture of Gaussians for a continuous design space, and shift the distribution towards high-performing designs in a manner analogous to a policy gradient update. In this work, we assume that the design space is discrete and that the number of designs is practically enumerable, and thus employ a categorical distribution.
Following the principle of maximum entropy, we model $p(\omega)$ as a Gibbs distribution:
\begin{equation}
    p(\omega) = \frac{e^{\beta R_\omega}}{\sum_{\omega \in \Omega} e^{\beta R_\omega}}
\end{equation}
where $R_\omega$ is the most recent reward obtained by design $\omega$, and $\beta$ is an inverse temperature parameter used to control entropy.
At each point in training, we set $\beta$ to maintain a decaying entropy target.
Specifically, we set $\beta = 0$ to specify a uniform distribution for an initial training period, and then decay entropy according to a linear schedule.
This schedule is akin to removing a constant fraction of designs from the search space at each step during training.

\begin{algorithm}[!t] \label{alg:main}
  \caption{Joint Optimization of Design and Control} \label{algorithm}
  \begin{algorithmic}[1]
      \STATE Initialize $\pi_\theta(a \vert s,\omega)$, $p(\omega)$, $T=0$
      \WHILE{True}
        \STATE Sample designs $\omega_1, \omega_2, \hdots, \omega_n \sim p_\phi$ \label{alg:sample}
        \STATE Control $\omega_1, \omega_2, \hdots, \omega_n$ with $\pi_\theta$ for $t$ timesteps. Add transitions to replay buffer. \label{alg:control}
        \STATE Update $\theta$ using soft actor-critic. \label{alg:sac}
        \STATE Update $R_{\omega_1}, R_{\omega_2}, \hdots, R_{\omega_n}$ with their obtained returns.
        \STATE Set timestep $T = T + nt$
        \STATE Set $\beta_T$ to match entropy target $\mathcal{H}_T$.
        \STATE Set $p(\omega) = \frac{e^{\beta_T R_\omega}}{\sum_\Omega e^{\beta_T R_\omega}}$
      \ENDWHILE
  \end{algorithmic}
\end{algorithm}

\subsection{Application to Soft, Legged Robots}

\begin{figure}
    \centering
    \begin{minipage}{0.475\linewidth}
        \subfigure[Design Space]{\includegraphics[width=\linewidth]{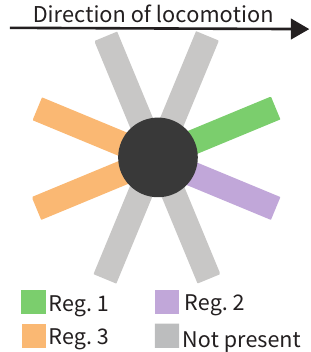}\label{fig:design_space_simple}}
    \end{minipage}\hfil
    \begin{minipage}{0.5\linewidth}
        \subfigure[Simulated Design]{\includegraphics[width=\linewidth]{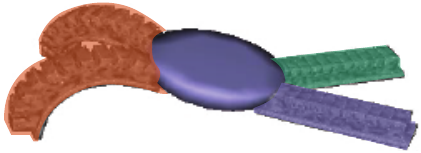}\label{fig:design_space_sim}}\\
        \subfigure[Real-world Design]{\includegraphics[width=\linewidth]{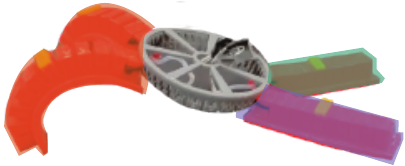}\label{fig:design_space_real}}
    \end{minipage}
    \caption{A visualization of \subref{fig:design_space_simple} our design space that consists of a disk with $N=8$ candidate locations for pneumatic actuators, each of which can be connected to one of $M=3$ pressure regulators. On the right are examples of a \subref{fig:design_space_sim} simulated and \subref{fig:design_space_real} real-world design, where colors denote the pressure regulator for each actuator. The forward direction is to the right.} \label{fig:design_space}
\end{figure}
The design space that we study here (Fig.~\ref{fig:design_space}) consists of a disk with $N$ equally spaced positions where soft, pneumatic actuators can be positioned radially outward. Each actuator is connected to one of $M$ different pressure regulators. Designing the robot then amounts to choosing whether (or not) to place an actuator at each of the $N$ locations and, for each placed actuator, connecting it to one of the $M$ pressure regulators. %
Our specific implementation considers
$N=8$ candidate locations and $M=3$ regulators, and restricts the design to having between three and six actuators. This results in a total of $41202$ unique designs, which can be reduced to $6972$ by exploiting symmetry in the regulator assignments. Each actuator is a PneuNet~\cite{mosadegh2014pneumatic} which, like similar soft actuators, has been combined to achieve crawling gaits~\cite{Goury18, shepherd_multigait_2011,gamus2020understanding,vikas2016design}, providing a well-studied baseline.

While our approach is compatible with any RL algorithm, we use the standard soft actor-critic (SAC) algorithm because it offers stable and data-efficient learning dynamics.
We train an open-loop controller modeled as a feed-forward neural network for the task.
This simplification of the controller forces the design to perform ``morphological computation''~\cite{rus2015design} to enable intelligent behavior.
The policy takes as input the design parameters along with the four most-recent actions and outputs pressure targets for each regulator. 
\section{Design-Reconfigurable Model Order Reduction} \label{sec:mor}

To obtain a finite-element model (FEM) whose speed is tractable for reinforcement learning, we perform model-order reduction (MOR).
This section presents our extension of the MOR method proposed by \citet{Goury18} to the problem of co-optimization.

\subsection{Reduction through Snapshot POD and Hyperreduction}\label{sec:subsec:mor_overview}

The finite element method provides an approximate numerical solution to partial differential equations (PDEs) by discretizing space into a mesh consisting of a set of finite elements.
Often, dense meshes (and subsequently, large amounts of computation) are needed to reach acceptable accuracy.

The soft actuator mesh contains nodes with position $q_{t_n}$ and velocity $v_{t_n}$ at discrete time step $t_n$. At each $t_n$, simulation requires solving a discrete form of Newton's second law~\cite{Goury18}:

\begin{equation} \label{eq:fem}
    A(q_{t_n}, v_{t_n})dv = b(q_{t_n}, v_{t_n}) + H^\top \lambda,
\end{equation}
where $dv = v_{t_{n+1}} - v_{t_n}$, $A \in \mathbb{R}^{d \times d}$ collects inertial and internal forces, $b \in \mathbb{R}^d$ contains terms from internal and external forces, and $H^T\lambda \in \mathbb{R}^d$ collects constraints (e.g., associated with contact with the floor), with $d$ being the number of degrees-of-freedom in the mesh.
When using dense meshes for accurate simulation, constructing the matrix $A$ and solving this system of equations are often the main bottleneck in FEM simulations.

We first reduce the system dimension through snapshot proper orthogonal decomposition (POD). Using the methods of ~\citet{Goury18}, we find a low-dimensional subspace $\Phi$ that well-approximates the space of possible motions and deformations while reducing the order through a Galerkin projection onto Equation~\ref{eq:fem}:%
\begin{equation} \label{eq:galerkin}
    \Phi^\top A(q_{t_n}, v_{t_n})\Phi d\alpha = \Phi^\top b(q_{t_n}, v_{t_n}) + \Phi^T H^\top \lambda.
\end{equation}
We achieve this by recording ``snapshots'' of the position $q_t$ of the mesh throughout a series of predefined motions that try to cover the space of common deformations.
The simulation then uses lower-dimensional coordinates $\alpha$, with $q_{t_n} = q_0 + \Phi\alpha_{t_n}$.

Though snapshot POD reduces the time to solve Equation~\ref{eq:fem}, it still requires computing the high-dimensional matrix $A$ at every time step. We therefore perform a hyperreduction to further approximate $A$ by predicting its entries from the contributions of a small number of elements. We use the hyperreduction method of energy conservation sampling and weighting (ECSW)~\cite{Farhat14}. For further details regarding this two-part method and a demonstration in soft robotics, we refer the reader to \citet{Goury18}.

\subsection{Modularized Reduction for Design-Reconfigurability} 

\begin{figure}[t]
    \centering
    \includegraphics[width=\linewidth]{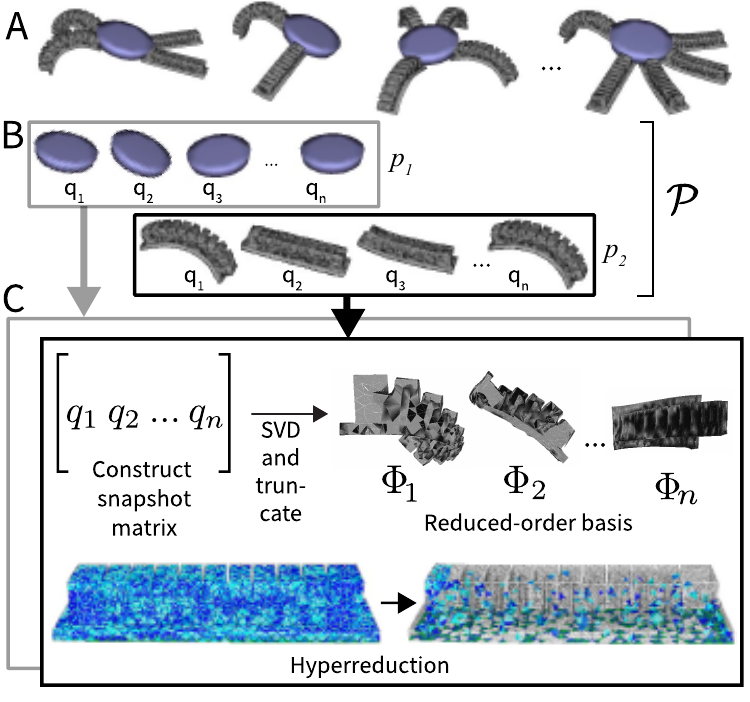}
    \caption{Our proposed technique for model order reduction that is compatible with co-optimization. \textbf{A}: Sample candidate designs and record their animations. \textbf{B}: Collect parts across designs and transform them into a common reference frame. \textbf{C}: For each part, perform a snapshot POD reduction and hyperreduction to obtain a reduced-order basis for motion, and reduced integration domain. We create reductions for new designs by combining reductions of their parts.} \label{fig:mor}
\end{figure}
The MOR method described in the previous section uses a single fixed mesh.
For high-dimensional soft robot design spaces, separately reducing each design is computationally intractable.
Instead, we define a modularized design space: each design $\omega \in \Omega$ is defined as a combination of a small set of fixed parts $\mathcal{P}$.
We then reduce each part in $\mathcal{P}$ using the method of Section \ref{sec:subsec:mor_overview} independently and combine the parts in arbitrary ways to form new designs. The number of times we perform MOR is then of the same size as $\mathcal{P}$ rather than the size of $\Omega$.

MOR on the modularized design space only well-approximates the full-order model when the computed subspace $\Phi^p$ for each part $p \in \mathcal{P}$ is close to all frequently achieved deformations.
Because designs will deform in different ways, it is necessary to include `snapshots' of motions from a large set of designs to achieve high-quality reduced-order models.
Therefore, careful snapshot selection on each module in $\mathcal{P}$ is crucial and inaccuracies may be exploited during optimization to result in invalid design-control pairs.
We achieve high-quality reduced-order models for each part by collecting snapshots from a heuristically chosen subset of $\Omega$ and animating those designs by cycling through the pressure extremes of each actuator.
When constructing the reduced basis for new designs, we transform the basis $\Phi_p$ to match the initial pose $(t_i, R_i)$ of each part $p_i$ by rigidly rotating the node positions that make up each basis vector:
\begin{equation}
    \Phi^{p_i}_j = \begin{bmatrix}
        R_i \Phi^p_j[0:3] & R_i \Phi^p_j[3:6] \cdots R_i \Phi^p_j[n-3:n]%
        \end{bmatrix},
\end{equation}
where $\Phi^p_j \in \mathbb{R}^n$ is the $j^\textrm{th}$ column of $\Phi^p$ and $\Phi^p_j[k:l]$ is a slice of that vector from index $k$ to index $l$.
We ignore the initial translation $t_i$ because translation basis vectors are included in $\Phi_p$. Figure~\ref{fig:mor} gives an overview of this approach.

\subsection{Reduction of Crawling Soft Robots}

We apply this reduction technique to our design space of crawling soft robots.
Our designs are composed of two parts: the central disk, and some number of identical PneuNets.
Therefore, the above approach allows us to perform two reductions (one for each part) as opposed to reducing each of the $6972$ designs in our design space.
We found that a sparse disk mesh was sufficiently fast and accurate for simulation and we therefore only reduce the PneuNet.

For the reduction of the PneuNet, we select a heuristic set of $256$ designs for which we collect snapshots.
Each design contains a unique subset of the eight potential PneuNet positions, and each PneuNet is controlled independently.
Similar to \citet{Goury18}, we iterate through the extremes of each actuator and record snapshots at fixed time intervals.
This can be seen as a walk through the vertices of an $n$-dimensional hypercube, where $n$ is the number of PneuNets present.
In order to verify the accuracy of our reduction, we evaluate it on a set of four test designs and animations by computing the distance between the node positions of the reduced and unreduced models. 
We perform a grid search over the two tolerances in the reduction algorithm and select the reduction that has the best time-accuracy trade off. See Appendix~\ref{app:mor} for more details. %
\section{Experiments} \label{sec:expt}
\begin{figure}[!t]
    \centering
    \subfigure[Baseline Design]{\includegraphics[height=1.75in]{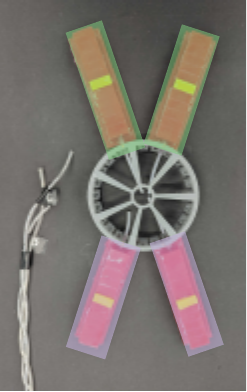}\label{fig:baseline_design}}\hfil
    \subfigure[Experimental Setup]{\includegraphics[height=1.75in]{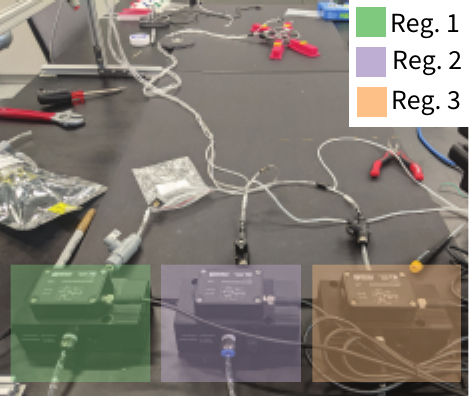}\label{fig:baseline_setup}}\\
    \subfigure[Baseline Gait]{\includegraphics[width=\linewidth]{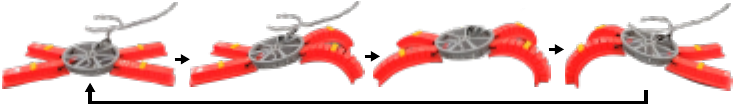}\label{fig:baseline_gait}}\\
    \subfigure[Reward vs.\ Phase]{\includegraphics[width=\linewidth]{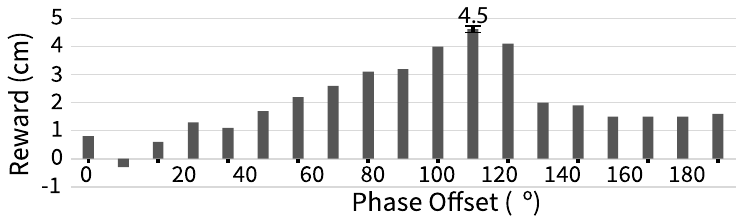}\label{fig:baseline_phase}}
    \caption{Our baseline consists of \subref{fig:baseline_design} an expert-designed soft robot with four legs, where the fore and hind legs are attached to pressure regulators one and two, respectively. The experimental setup consists of \subref{fig:baseline_setup} three pressure regulators, a crawling surface, and 3.15\,mm outer-diameter %
    tubing connected to the robot (in the the distance). \subref{fig:baseline_gait} Snapshots of the expert-designed gait. \subref{fig:baseline_phase} Reward (distance traveled) obtained by an offset-sine gait with different phase shifts. A phase difference of $110^{\circ}$ achieves the highest reward.}
    \label{fig:gait_tuning}
\end{figure}
We test our approach by attempting to find a design and open-loop controller that crawl as far as possible on a flat plane in a $20$ second episode.
We define reward as the distance traveled in the (forward) $x$-direction (in cm) as measured at the center of the disk.

After performing model order reduction, we carry out FEM simulation using the SOFA Finite Element framework~\cite{Faure2012} with the soft robotics~\cite{Coevoet17} and model order reduction~\cite{Goury18} plugins.
We model the PneuNet legs (including the inflatable and constraint material) and central disk as linear elastic materials.
We assign input pressures as pressure constraints on surface meshes internal to the PneuNets.
We estimated the Young's modulus of the PneuNet material (Smooth-On DragonSkin 30) based on the published Shore hardness together with the method of~\citet{qi2003durometer}.
We modeled the constraint layer of the PneuNet as being linear elastic with a Young's modulus twice the magnitude of the inflatable material.
We tuned the Poisson's ratio in order to maintain numeric convergence and qualitative realism.
The friction model used is Coulomb friction.
To account for any inaccurate or unmodelled effects, we measure deformation of a single, real PneuNet under fixed pressures, find the corresponding pressures that results in the same deformation of the simulated PneuNet, and fit an affine function to this data.
Pressures commanded by our learned policies are then mapped through this function to ensure a simulated response similar to that of the real PneuNets.
We find that this step greatly improves sim-to-real transfer.

We built an experimental platform to evaluate the performance of our learned soft robot designs. The platform consists of a pressure chamber at 400$\pm100$ kPa in series with three pressure regulators. %
Each pressure regulator is connected by a lightweight tube to the robot and is controlled by a programmable power supply. %
We created a modular assembly scheme in which any robot from the design space can be built. We 3D-printed a lightweight polymer disk that was then attached to molded, soft PneuNet legs~\cite{mosadegh2014pneumatic} (Smooth-On DragonSkin 30). Appendix~\ref{app:robotbuild} provides additional details on the fabrication procedure.
We designed a baseline design-controller pair similar to the robot used by \citet{shepherd_multigait_2011}.
The baseline (Fig.~\ref{fig:gait_tuning}) has two fore legs and two hind legs placed $45^\circ$ apart with each pair controlled by a single regulator.
Based on recent analysis of inching gaits~\cite{gamus2020understanding}, we constrain each pressure regulator to produce a sine wave of equal amplitude and period.
We achieve forward motion by imposing a phase shift between the sine waves for the fore and hind legs.
We select a pressure range of $0$ to $90$\,kPa to avoid both physical instabilities (i.e., aneurysms of the PneuNets) and numerical instabilities in the FE simulator.
We use the maximum amplitude allowed in this range of $45$\,kPa and choose a period of $4$\,sec, which is the fastest period that led to stable motion.
The optimal phase shift depends heavily on friction~\cite{gamus2020understanding} so we conducted experiments with different phase shifts between $0^{\circ}$ and $180^{\circ}$ in increments of $10^{\circ}$, and chose the value that resulted in the highest reward. Figure~\ref{fig:gait_tuning} shows the effect of phase shift on the reward.
We use $96$ parallel environments for data collection.
Each environment contains a design sampled from $p(\omega)$ (Algorithm~\ref{alg:main}, line~\ref{alg:sample}) that is controlled with the current policy for one episode (Algorithm~\ref{alg:main}, line~\ref{alg:control}).
The control policy is then updated using the soft actor-critic algorithm on data from a replay buffer (Algorithm~\ref{alg:main}, line~\ref{alg:sac}).
We repeat this process for $1$M environment timesteps during which we fix the design distribution to be uniform for the first $200$K timesteps, after which we linearly decay the entropy to zero at $1$M timesteps. 
See Appendix~\ref{app:params} for more details on the experimental setup, including a full list of simulation and learning parameters. After training, we manufacture several of the highest-performing designs and conduct a series of experiments to evaluate the sim-to-real transfer of the learned design-control pairs.
\begin{figure}[!t]
    \centering
    \includegraphics[width=\linewidth]{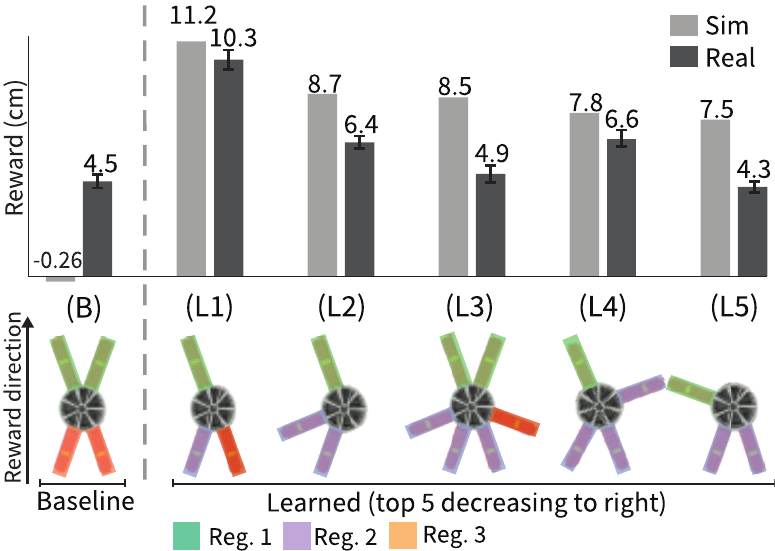}
    \caption{A comparison of the reward (distance traveled in cm) achieved in simulation and reality for the duration of the $20$\,sec episode for the baseline robot (B) and the top-five learned design-control pairs (L1--L5).}\label{fig:comparison_reward}
\end{figure}

\section{Results}\label{sec:results}

We examined the performance of the top-five learned design-control pairs that our framework discovers, our baseline design-control pair, and their capacity for sim-to-real transfer.
We refer to the baseline as ``B'' and the learned pairs as ``L1--L5'' in order of decreasing reward.
Figure~\ref{fig:comparison_reward} visualizes these designs and compares their reward to that of the baseline in simulation and in the real world (tested over five trials).
The top four learned robots (L1--L4) outperform the baseline in both simulation and the real world, while the fifth robot (L5) performs comparably to the baseline in the physical experiments. The top learned robot, L1 outperforms the baseline (in real measurements) by a factor of $2.3$.

\begin{figure}[!t]
    \centering
    \includegraphics[width=\linewidth]{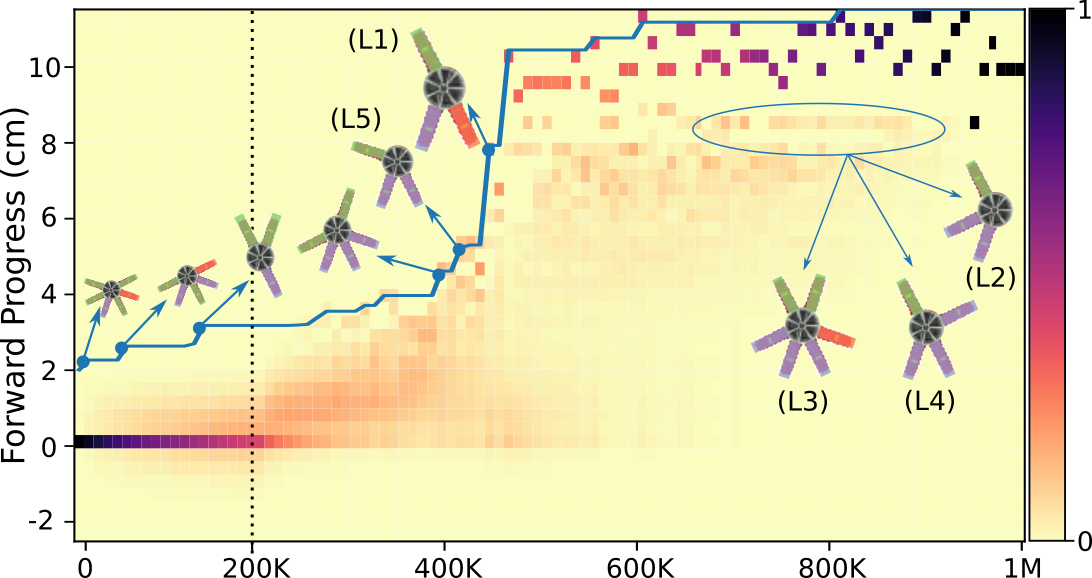}
    \caption{A histogram of rewards throughout training, weighted by the design distribution. The blue line represents the highest reward achieved at each point throughout training. The designs pictured on the left show important mode switches, and the right shows other high-performing designs. The design distribution is fixed to be uniform until $200$K timesteps and then a temperature parameter is tuned to match a linearly decaying entropy target.}
    \label{fig:reward_hist}
\end{figure}

Figure~\ref{fig:reward_hist} inspects the training dynamics of our approach; it shows a histogram of rewards achieved by the design distribution throughout training.
In the beginning of training, the design distribution is constrained to be uniform and nearly every design achieves zero reward.
Starting at $200$K timesteps, the algorithm constrains the distribution with a linearly decaying entropy, after which the algorithm specializes to high performing designs (i.e., L5).
Approximately halfway through training, the algorithm converges on design L1, which achieves a reward that is several centimeters better than the next-best design-control pair.

\subsection{Qualitative Analysis of Learned Designs and Gaits}

\begin{figure*}[t!]
    \centering
    \subfigure[Baseline Policy]{\includegraphics[width=0.51\linewidth]{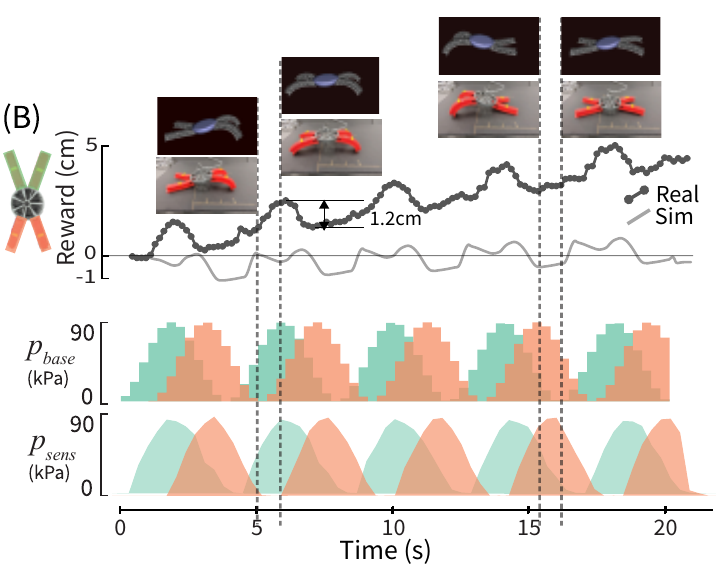}\label{fig:comparison_policy_base}}%
    \subfigure[Learned Policy with Highest Reward]{\includegraphics[width=0.51\linewidth]{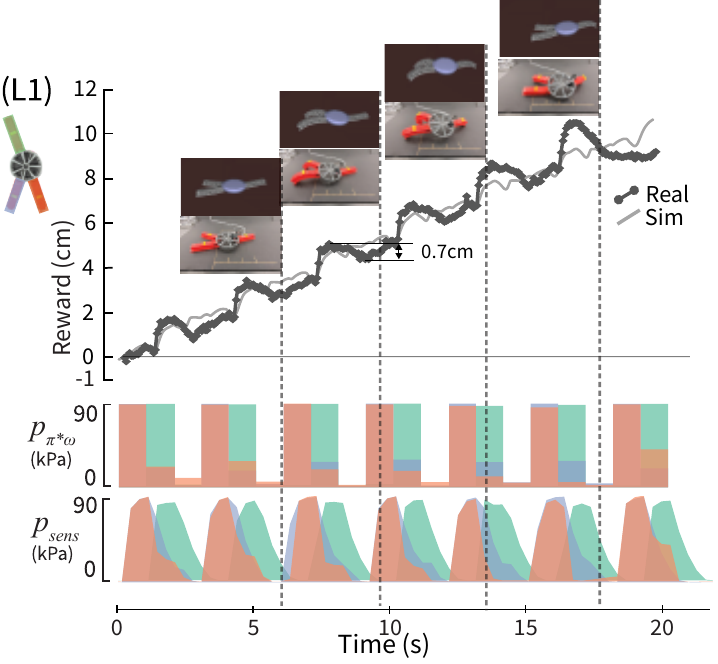}\label{fig:comparison_policy_learned}}
    \caption{A comparison between the \subref{fig:comparison_policy_base} baseline (B) and \subref{fig:comparison_policy_learned} highest-performing learned robot (L1) in terms of the reward (distance traveled in cm) achieved in simulation and reality for the duration of the $20$\,sec episode, along with the corresponding control policy in terms of commanded and sensed pressures.}\label{fig:comparison_policy}
\end{figure*}

Observing the learned design-control pairs in simulation and through real-world experiments reveals that they exploit changes in frictional forces in clever ways to create forward motion. Figure~\ref{fig:comparison_policy} compares the open-loop (pressure) gaits and reward trajectories (in simulation and the real-world) over the duration of the $20$\,sec episode for the baseline and top learned design-control pairs. The baseline robot (B) uses a symmetric design that moves forward through out-of-phase actuation of the front and hind legs (Fig.~\ref{fig:comparison_policy_base}). In contrast, robot L1 uses an asymmetric design alongside out-of-phase actuation of the three attached pressure regulators. The result of this design and motion is a pivoting behavior visible in the photos within Figure~\ref{fig:comparison_policy_learned} and the supplementary video. Due to its asymmetry, the soft robot rolls to its side and the contact area of the front leg with the floor is reduced (leading to reduced frictional forces on that leg). This reduction in contact area enables the front PneuNet `leg' to slip forward instead of pushing backward. The result is that robot L1 uses forward-slipping motion without as much backward displacement per step (a.k.a.\ backsliding) in L1 ($0.7$\,cm backsliding) compared to the baseline ($1.2$\,cm backsliding).

\begin{figure}[t]
    \centering
    \includegraphics[width=1.0\linewidth]{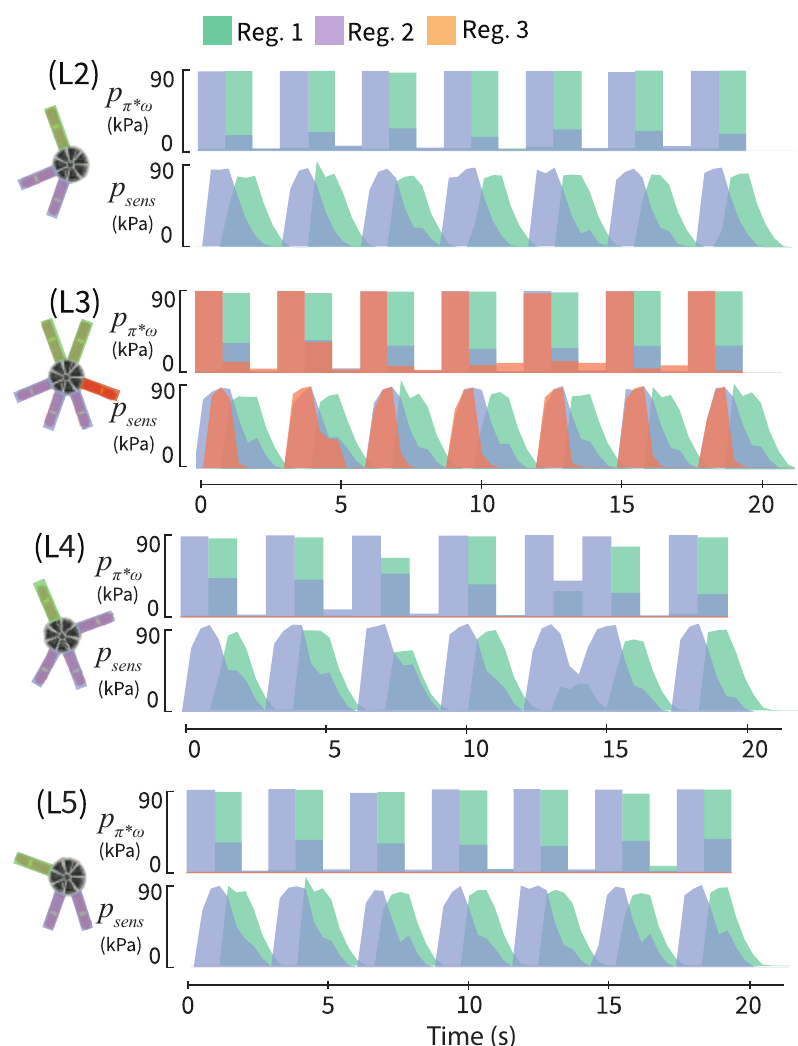}
    \caption{Commanded and sensed pressures $p_{\pi*\omega}$ and $p_\textrm{sens}$ for the second- to fifth-ranking learned design-controller pairs L2--L5.}
    \label{fig:policy_L25}
\end{figure}

Figures~\ref{fig:reward_L25} and \ref{fig:policy_L25} show the reward per timestep and pressurization (commanded and sensed) for robots L2--L5.
Four of the five top robots (L1, L2, L4, and L5) also leverage asymmetric morphologies (Fig.~\ref{fig:comparison_reward}) to tilt the front leg onto its corner or edge and reduce friction by reducing contact. Gaits for the L2--L5 designs (Fig.~\ref{fig:policy_L25}) operate in similar stages to L1: the hind legs inflate first to anchor the robot and subsequently follow a cycle that is out of-phase with the cycle followed by the front legs. While the morphology of robot L3 is symmetric, the use of regulator three on only one leg adds asymmetry to its gait.
Interestingly, each of the learned gaits inflate the hind legs before the fore legs and the baseline does the opposite. We find that the learned gaits, especially L1, L2, and L5, make the most forward progress at the transition between inflation of the fore legs and deflation of the back legs (e.g., overlap of green and orange sensed pressures in Figure~\ref{fig:comparison_policy_learned}). We hypothesize that the asymmetry of the learned designs makes this motion possible.
See our webpage\footnote{\href{https://sites.google.com/ttic.edu/evolving-soft-robots}{https://sites.google.com/ttic.edu/evolving-soft-robots}} for side-by-side recordings of the learned designs and gaits in simulation and reality. %

\subsection{Sim-to-Real Transfer}

Given the goal of being able to co-optimize the design and control of physically realizable soft robots, we compare the learned robots and baseline in physical experiments.
We measure the \textit{zero-shot} sim-to-real transfer performance by manufacturing the learned designs and applying the learned policy without modification. %

We find that all of optimized designs and gaits make consistent forward progress, with four out of five (L1--4) outperforming the baseline in physical measurements, and one performing comparably (L5) to the baseline.
The highest-performing robot (L1) and robot L4 have strong agreement with simulation; with the standard deviation across trials taken into account, real reward reaches within 1\,cm of the simulated reward.

\begin{figure}[t]
    \centering
    \includegraphics[width=\linewidth]{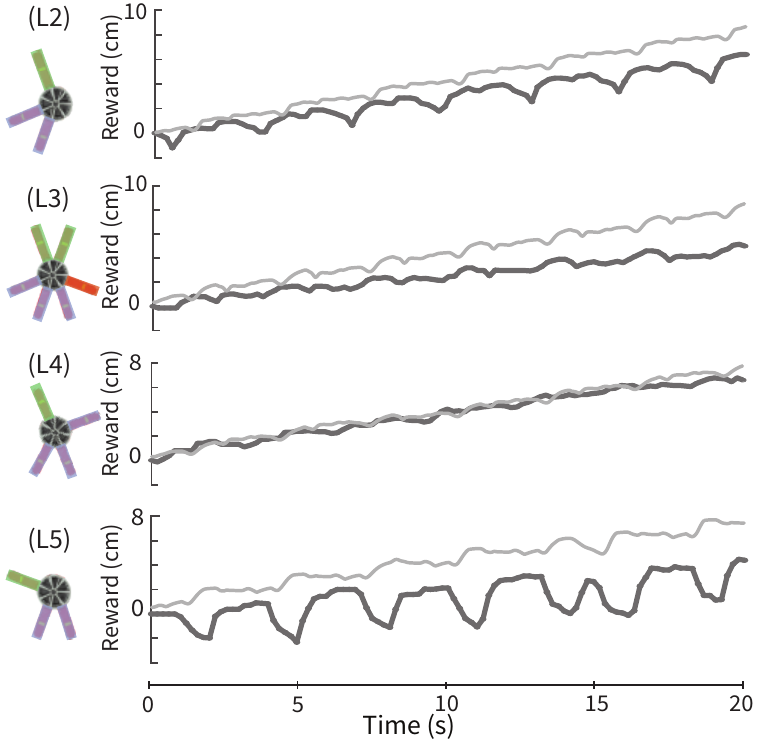}
    \caption{Reward per timestep achieved in simulation and reality by the second- to fifth-ranking learned design-controller pairs L2--L5.}
    \label{fig:reward_L25}
\end{figure}

Despite strong qualitative shape agreement seen in the supplementary video, L2, L3, and L5 see a pronounced drop in performance. A comparison of the simulated and real reward over time shows the main reason for this performance decrease.
Figure~\ref{fig:reward_L25} reveals that while these reward functions have qualitatively similar waveforms, the backsliding phases of the waveforms appear to have higher magnitude in reality than in simulation.
The initial phase of each of these gaits inflates the back legs, relying on the friction from the front leg(s) to hold the robot in place rather than slide backwards.
The front PneuNets slip very little in simulation, but they maintain less grip in the real world. The resulting error in the backsliding phase of each of these gaits is then accumulated with every step, resulting in less forward motion.

Figure~\ref{fig:comparison_policy_base} compares the forward progress of the simulated baseline against its real-world counterpart.
Despite detailed tuning of simulation parameters, our baseline gait achieves poor reward in simulation.
The simulated baseline gait does indeed make slow forward progress. Yet, in key segments of the gait (shown by images in Fig.~\ref{fig:comparison_policy_base}), the simulation records less progress, or backward progress, compared to the real-life measurement.
As shown by prior analysis of crawling robots~\cite{gamus2020understanding,vikas2016design}, this type of sinusoidal gait is very sensitive to frictional forces.
Because the design is symmetric, the gait relies on the subtle differences in friction to enable timing of stick-slip interactions for forward motion.
As a result, error accrual due to backward sliding that was already evident in the learned designs is of even higher magnitude here.

While there are many other differences between simulation and reality, such as dynamics associated with the pressure regulators and damping of high frequency motions, we find that the modelling errors associated with stick-slip transitions the largest effect on the transfer performance of our designs.
Stick-slip transitions are notoriously difficult to model~\cite{luo15}, and often requires smoothness approximations for numeric stability. Overall we still see mostly successful transfer for all five optimized designs.
\section{Conclusion} \label{sec:conclusion}

This work describes a complete framework for the simulation and co-optimization of the design and control of soft robots capable of zero-shot sim-to-real transfer. We present an algorithm for co-optimization and a framework to create reconfigurable, reduced-order models for soft robotics. Experiments demonstrate that our framework learns design-gait pairs that outperform an expert-designed baseline in a soft robot locomotion task. We further characterize the successful qualitative and quantitative transfer of these learned pairs from simulation to reality.

Soft robots are compliant-bodied and mechanically intelligent. As a result, their design and control spaces are not well-separated.
For many tasks it difficult or impossible to explore these design/control spaces through prototypes or analysis alone; simulation is needed to build and evaluate designs and controllers in a tractable manner.
Our work takes a step in this direction by showing that it is possible to combine reinforcement learning techniques with finite element simulation to deliver fast and physically accurate co-optimization for soft robotics.

This work has some limitations.
Modelling errors, due to friction, linear elasticity, etc., caused us to linearly warp simulation parameters to obtain numerical stability and realism.
Even so, these errors led to a degradation of transfer performance for some designs.
Future work will include improved simulation techniques as well as investigations into domain adaptation and domain randomization methods as a means of improving sim-to-real transfer.
Given the success of this method for simple locomotion, we plan to explore adaptations of this framework to different design and control spaces as well as more complex tasks such as manipulation.%

\section*{Acknowledgments}
We thank Olivier Goury and Hugo Talbot for assistance implementing SOFA and the MOR module, and Arthur MacKeith for additional simulation support.

\printbibliography
\flushcolsend

\clearpage
\appendices

\section{Learning and Simulation Parameters} \label{app:params}

Table~\ref{tab:sac_param} lists the hyperparameter settings that we used for co-optimization.
We represent each design as a concatenation of eight four-dimensional one-hot vectors indicating whether each PneuNet is present and which regulator it is connected to.
Our policy is open-loop and receives the last four pressure actions as input.
The actions are concatenated with the design representation to create a $44$-dimensional input to the policy and value networks.
Both the policy and value networks are four-layer feed-forward networks and each hidden layer has $[400, 200, 100]$ units, respectively.
We train on a $32$-core AMD EPYC $7502$, with the experiment taking eight days to complete.
\begin{table}[h]
    \centering
    \begin{tabularx}{\linewidth}{Xr}
        \toprule
        Hyperparameter  &  Value \\
        \midrule
        Number of environments & $96$ \\
        Maximum timesteps & $1$M \\
        Buffer size & $100$K \\
        Batch size & $512$ \\
        Discount factor ($\gamma$) & $0.95$\\
        Policy learning rate & $0.0006$ \\
        Q-function learning rate & $0.0020$ \\
        Optimizer & $\textrm{ADAM}(\beta_1 = 0.9, \beta_2 = 0.999)$ \\
        Target network smoothing coefficient & $0.01$ \\
        Learning start & $10$K \\
        Timesteps per SAC update & $1$ \\
        Timesteps per sampled design & $20$ (one episode) \\
        Entropy linear decay start & $200$K \\
        Entropy linear decay end & $950$K \\
        Entropy target start & uniform \\
        Entropy target end & $0$ \\
        \bottomrule
    \end{tabularx}
    \caption{Soft Actor Critic and Co-optimization Parameters}
    \label{tab:sac_param}
\end{table}

Table~\ref{tab:sim_param} provides our simulation parameters.
We use a linear elastic model and set Young's modulus for the silicone using known material parameters.
The Young's modulus for the paper and disk were chosen to be significantly stiffer that the silicone.
The Poisson ratios were set to maintain realism while avoiding numerical instabilities.
The masses were measured from their real counterparts but then scaled by a factor of $2.5$ to avoid numerical instability.
The friction coefficient was chosen to qualitatively approximate the stick-slip behavior of the baseline design when all the PneuNets are inflated and deflated simultaneously.

\begin{table}[!h]
    \centering
    \begin{tabularx}{\linewidth}{Xr}
        \toprule
        Parameter & Value \\
        \midrule
        Elastic model & linear \\
        Young's modulus for PneuNet silicone & $1160$\,kPa\\
        Poisson ratio for PneuNet silicon & $0.2$ \\
        Young's modulus for PneuNet paper & $2320$\,kPa\\
        Poisson ratio for PneuNet paper & $0.49$ \\
        Young's modulus for disk & $5000$\,kPa \\
        Poisson ratio for disk & $0.3$ \\
        PneuNet mass & $105$\,g \\
        Disk mass & $75$ g \\
        Friction coefficient & $1.2$ \\
        Gravity & $9800$\,mN\\
        \bottomrule
    \end{tabularx}
    \caption{SOFA simulation parameters}
    \label{tab:sim_param}
\end{table}

\begin{figure}[!h]
    \centering
    \includegraphics[width=\columnwidth]{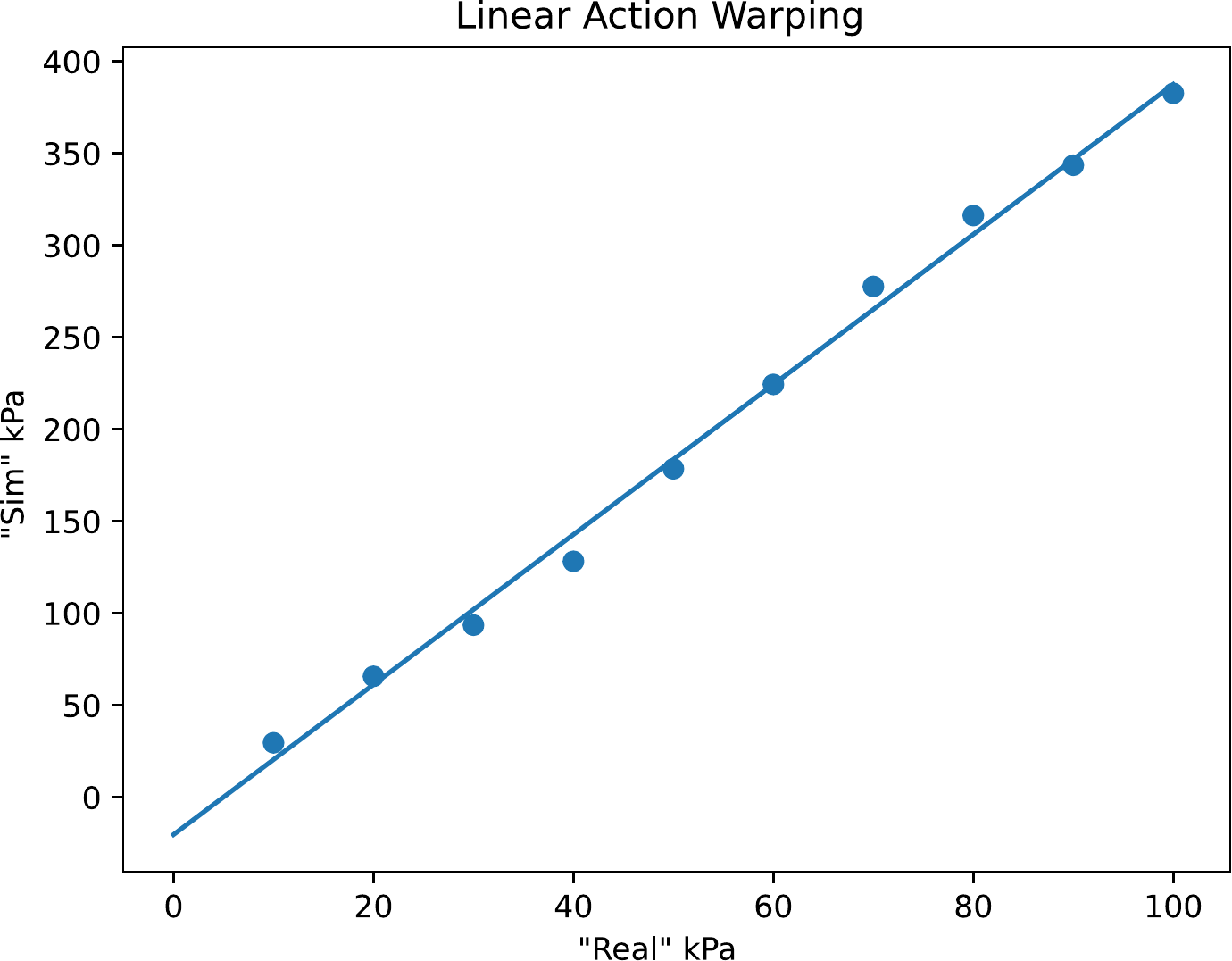}
    \caption{The linear pressure scaling used to align the behavior of a single PneuNet in simulation and reality. In simulation, we scale the pressures output by the policy prior to applying we apply them. This helps to reduce discrepancies between simulation and reality caused by modelling error.}
    \label{fig:warping}
\end{figure}

Figure \ref{fig:warping} shows the linear pressure scaling used to align the PneuNet bending response between simulation and reality.
To calibrate the action scaling, we record the bending of a single PneuNet under pressures from $10$\,kPa to $100$\,kPa, in increments of $10$\,kPa.
We then search for the pressures in simulation which achieve an equivalent bend and fit a linear function to the results.

\vspace{10ex}

\section{Model Order Reduction Evaluation} \label{app:mor}

\begin{figure*}[!t]
    \centering
    \includegraphics[width=\textwidth]{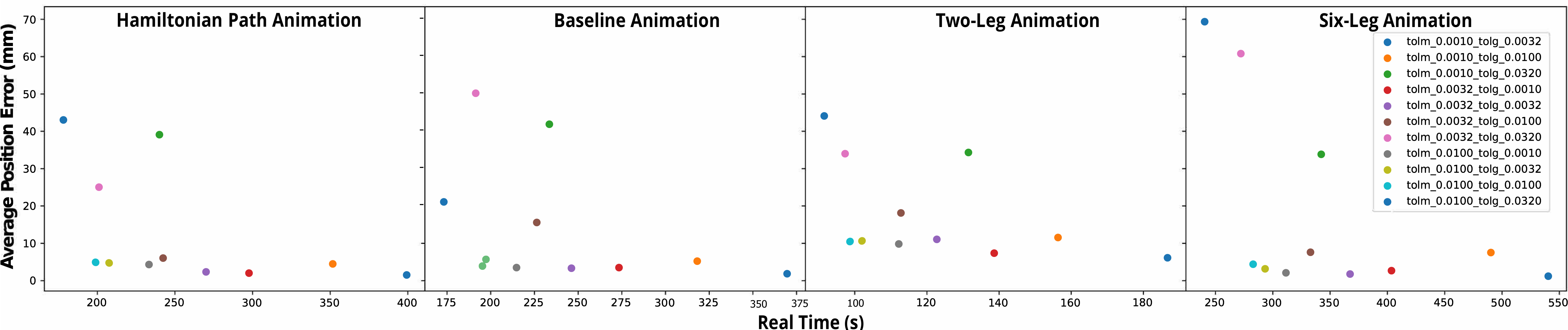}
    \caption{This plot shows the trade off between positional errors introduced by the reduction and the time to simulate each animation for a $3$x$3$ grid search over the $2$ MOR tolerances. We choose a mode tolerance of $0.0032$ and a hyperreduction tolerance of $0.0010$ (red dot) for our experiments.}
    \label{fig:mor_errors}
\end{figure*}

The snapshot-POD reduction method has two tolerances that must be set, one for the error induced by the basis $\Phi$, and one for the error induced by the hyperreduction. To explore the effect of these two parameters and to verify the validity of our reduced models, create a test set of four designs and animations:
\begin{enumerate}
    \item the \textbf{Hamiltonian animation} uses the baseline four-legged design and iterates through the pressure extremes of the four PneuNets.
    \item the \textbf{baseline animation} also uses the baseline design and a gait similar to the baseline gait from our experiments.
    \item the \textbf{two leg animation} uses a design with two alternately inflating PneuNets at a $90^\circ$ angle.
    \item the \textbf{six leg animation} uses a design with six PneuNets that alternately inflate in two groups of three.
\end{enumerate}
For each design and animation, we compare the node positions of the mesh using the unredeced and reduced-order models.
We perform a coarse grid search over the two MOR tolerances and plot the simulation speed and accuracy of each reduction.
Errors are computed as the L2 distance between the node positions of the reduced and unreduced model averaged over each node in the mesh and each timestep.
Based on this analysis, we use a mode tolerance of $0.0032$ and a GIE tolerance of $0.001$ for our experiments.

\section{Hardware and Fabrication Details} \label{app:robotbuild}
\begin{figure}[h]
    \centering
    \includegraphics[width=0.5\textwidth]{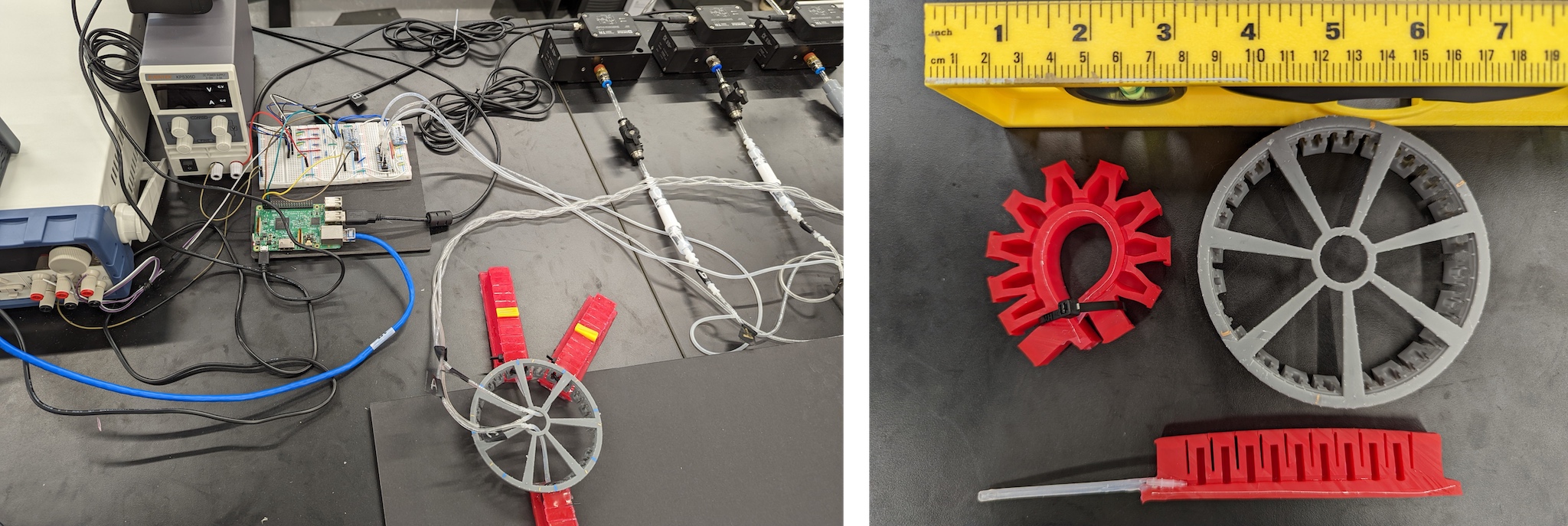}
    \caption{\textbf{Left}: A picture of our hardware set up consisting of a Raspberry Pi, three pressure regulators, two power supplies, three pressure sensor, and a breadboard connecting everything to the Pi. \textbf{Right}: A picture of our $3$D-printed disk and a molded PneuNet cut in half to display the internal structure.}
    \label{fig:robotbuild}
\end{figure}

Figure \ref{fig:robotbuild} displays our experimental platform and robot design.
We keep a pressure chamber at 400$\pm100$ kPa attached in series with three pressure regulators (Enfield TR-010-gs).
The learned and baseline policies are executed on a Raspberry Pi where the pressure commands are converted to voltage commands and sent to a programmable power supply (BK Precision 9129B).
The power supply then sends voltages to each of the three pressure regulators through a breadboard that converts the voltages into an acceptable range.
Each pressure regulator is connected by a lightweight tube to the robot.
We connect an external pressure sensor to each regulator to verify that the correct pressure is applied.

We created a modular assembly scheme in which any robot from the design space can be built.
We 3D-printed a lightweight polymer disk (Fig.~\ref{fig:robotbuild} (right)) with uniformly distributed locations  where we can attach the PneuNet actuators while routing the pneumatic cables away from the robot (Fig.~\ref{fig:robotbuild} (left)). We 3D-printed moulds for the PneuNets based on a modification of the design given in the Soft Robotics Toolkit~\cite{holland2014soft} such that the two end prismatic segments are filled rather than hollow (Fig.~\ref{fig:robotbuild} (right)), which allows the first prismatic segment to be used as an attachment point to the disk. We fabricated the PneuNet actuators using Smooth-On DragonSkin 30 silicone. %
 
\end{document}